\documentclass[letterpaper, 10 pt, conference]{ieeeconf}  

\IEEEoverridecommandlockouts                              

\usepackage{graphics} 
\usepackage{epsfig} 
\usepackage{mathptmx} 
\usepackage{times} 
\usepackage{amsmath} 
\usepackage{amssymb}  
\usepackage{graphicx}
\usepackage[utf8]{inputenc}
\usepackage{multirow}
\usepackage{subcaption}
\usepackage{caption}
\usepackage{hyperref}
\usepackage{comment}

\title{\LARGE \bf Robotic Scene Segmentation with Memory Network for \\
Runtime Surgical Context Inference}

\author{ Zongyu Li$^{1}$, Ian Reyes$^{2}$, Homa Alemzadeh$^{1}$ 
\thanks{$^{1}$ Department of Electrical and Computer Engineering, University of Virginia, Charlottesville, VA 22903 USA.
$^{2}$ Ian Reyes was with the Department of Computer Science, University of Virginia, Charlottesville, VA 22903 USA. He is now with IBM.
        {\tt\small \{zl7qw, ir6mp, ha4d\}@virginia.edu} code available: \href{https://github.com/UVA-DSA/Runtime_RobScene_Seg_2Context}{\text{https://github.com/UVA-DSA/Runtime\_RobScene\_Seg\_2Context}}%
}}

\begin{document}
\bstctlcite{IEEEexample:BSTcontrol}
\maketitle  
\thispagestyle{empty}
\pagestyle{empty}

\begin{abstract}
 Surgical context inference has recently garnered significant attention in robot-assisted surgery as it can facilitate workflow analysis, skill assessment, and error detection. However, runtime context inference is challenging since it requires timely and accurate detection of the interactions among the tools and objects in the surgical scene based on the segmentation of video data. On the other hand, existing state-of-the-art video segmentation methods are often biased against infrequent classes and fail to provide temporal consistency for segmented masks. This can negatively impact the context inference and accurate detection of critical states.  In this study, we propose a solution to these challenges using a Space-Time Correspondence Network (STCN). STCN is a memory network that performs binary segmentation and minimizes the effects of class imbalance. The use of a memory bank in STCN allows for the utilization of past image and segmentation information, thereby ensuring consistency of the masks. Our experiments using the publicly-available JIGSAWS dataset demonstrate that STCN achieves superior segmentation performance for objects that are difficult to segment, such as needle and thread, and improves context inference compared to the state-of-the-art. We also demonstrate that segmentation and context inference can be performed at runtime without compromising performance.
\end{abstract}


\section{Introduction}
Robot-assisted surgery has transformed the field of minimally invasive surgery by allowing surgeons to operate with greater dexterity and precision and improving patient outcomes. Characterizing the interactions among the surgical instruments and important objects and anatomical structures within the surgical scene can provide context awareness~\cite{COMPASS}, which is crucial for various downstream tasks, such as cognitive assistance \cite{scheikl2021cooperative}, skill evaluation \cite{tao2012sparse, varadarajan2009data, zhang2020automatic,papp2022surgical} and error detection \cite{yasar2019context, yasar2020real, hutchinson2022analysis, Li2022error, inouye2022assessing}.

However, accurate detection of surgical context from video is a challenging task. Various deep learning methods \cite{nwoye2020recognition,nwoye2022rendezvous,li2022sirnet} have been proposed to infer tool tissue interactions from surgical videos. These black-box models, however, suffer from lack of transparency and dependency on large labeled datasets. Recently, \cite{ICRA23} proposed logic operations on the masks of different objects in the surgical scene to infer surgical context. This method provides interpretability and enables efficient integration of expert knowledge in a domain the data is usually limited. However, this method requires precise segmentation masks to detect interactions between objects and instruments, such as contact and hold. For surgical scene segmentation, multiclass segmentation methods which focus on classification of each pixel are commonly used. This means that each pixel is assigned one class label with the highest probability. 
However, these methods have primarily focused on identifying graspers and common objects in porcine procedures \cite{allan20192017, allan20202018,garcia2017real,bodenstedt2018comparative,shvets2018automatic,jin2019incorporating} and  can have difficulty identifying small objects such as needles and rarely used instruments due to class imbalance. Other approaches focus on thread segmentation with fine-tuning \cite{lu2020learning} or performing 3-D computation through a calibrated stereo camera system \cite{lu2019surgical}. 
Another challenge in the recent state-of-the-art models is to correctly identify segmentation masks when the image deviates from the common viewpoint (e.g. the bending of graspers) or when there are occlusions (e.g. the interactions between needle and graspers) \cite{chadebecq2022artificial}. This problem can be potentially solved by ensuring mask consistency for an instrument through time by incorporating a temporal prior \cite{jin2019incorporating, zhao2022trasetr}. 

The recent development of Space-Time Memory Networks (STM) \cite{oh2019video,cheng2021rethinking} have achieved top performance in the semi-supervised video object segmentation (VOS) tasks on benchmark dataset DAVIS 2017 \cite{pont20172017} and YouTubeVOS 2018 \cite{xu2018large}. For these models, a memory bank is created for each object and the query frames are matched to these banks to retrieve information. This method effectively reduces the effect of label imbalance through binary segmentation and ensures label consistency over time. STM models perform well on videos containing objects commonly appearing in everyday life in an offline manner. However, these models have not been applied for segmentation of multiple objects in robotic scenes and have never been examined if they can perform segmentation at runtime, limiting their potential applications in tasks such as error detection \cite{Li2022error,yasar2020real}.


In this paper, we adapt a lightweight STM, the Space-Time Correspondence Network (STCN) \cite{cheng2021rethinking}, by changing the first image/mask pair, batching the input frames and fine-tuning on the robotic surgical dataset  (JIGSAWS \cite{gao2014jhu}) to perform runtime surgical scene segmentation. We then use the segmentation masks to perform surgical context inference and show that improved segmentation performance can lead to more accurate context inference. 

Specifically, we make the following contributions.
\begin{itemize}
    \item Adapt the STCN in video object segmentation to perform \textit{runtime} surgical scene segmentation.
    \item Show the superior performance of the STCN model in comparison to the state-of-the art single frame models for segmenting surgical instruments. 
    \item Demonstrate that the STCN Network achieves good segmentation performance even if the first image/mask pair does not come from a prior frame of the video. 
    \item Show that more precise segmentation masks lead to improved context inference, in particular more accurate detection of the interactions/states of the objects/instruments that are hard to segment.
    \item Demonstrate that STCN segmentation and context inference can be performed within runtime constraints with minimal influence on the performance. 

\end{itemize}

\section{Related work}
\textbf{Semantic segmentation} involves classifying each pixel in an image into a specific object or background. Various robotics scene segmentation challenges \cite{allan20192017, allan20202018} have focused on the task of semantic segmentation. One of the most popular segmentation frameworks to perform semantic segmentation is the UNet structure \cite{ronneberger2015u,garcia2017real,bodenstedt2018comparative,shvets2018automatic,jin2019incorporating}. But the deep learning models for semantic segmentation can suffer from the label imbalance problem where the model can be biased against small objects.

\textbf{Instance segmentation} involves detecting the presence of objects of interest in an image and segmenting each object instance from the background. By first identifying the instrument candidates and then assigning a unique category to them, the Mask R-CNN based methods focus on providing a binary mask for each specific type of instrument and could be a good solution to address the data imbalance problem.
In recent works, Mask R-CNN was adapted to perform fine-grained instrument segmentation \cite{gonzalez2020isinet,sun2022parallel}.  However, instance segmentation models perform instance segmentation on a specific frame and do not consider the evolution of the masks through time, resulting in inconsistent labels.

\textbf{Semi-supervised video object segmentation} tasks \cite{pont20172017} focus on estimating object masks in all video frames given the ground truth mask of the target object in the first frame. Space-Time Memory Networks (STM) are the top-performing models on challenge datasets such as DAVIS 2017 \cite{pont20172017} and YouTubeVOS 2018 \cite{xu2018large}. 
The STM performs binary segmentation by decoding the memory readout and integrates prior image and mask information to segment objects of the current frame. This effectively eliminates the need to perform multi-class classification on the pixel and improves temporal consistency of the masks. However, STM has been primarily used for offline video segmentation and has not been applied for \textit{runtime surgical scene segmentation}. 

\begin{figure*}[ht!]
    \centering
    \vspace{0.75in}
    \includegraphics[trim = 3in 4in 4.5in 2in,width=0.4\textwidth,height=4.2cm]{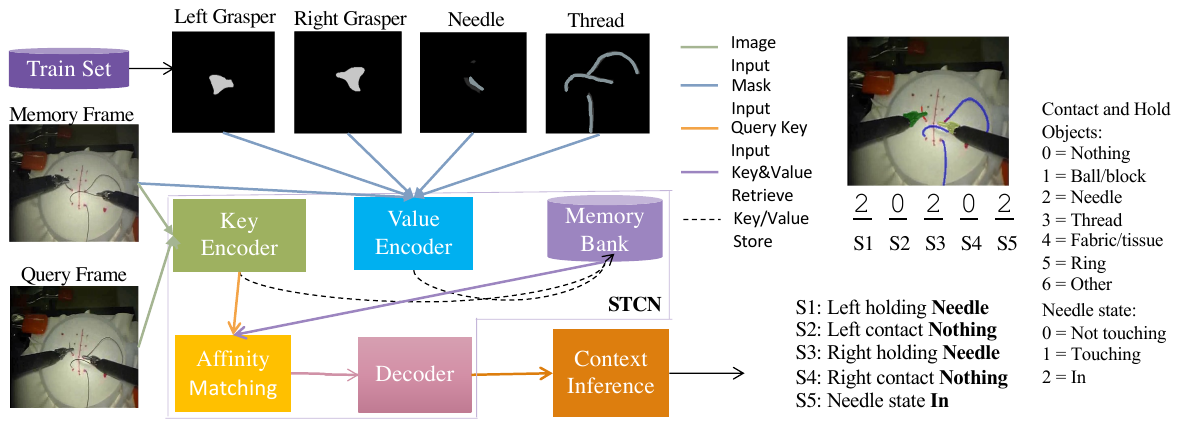}
    \vspace{-1em}
    \caption{Pipeline for Gesture Segmentation and Context Inference with STCN (Train-FF) Setup}
    \label{fig:pipeline}
    \vspace{-1em}
\end{figure*}

\textbf{Surgical context inference} focuses on detecting the values of a set of state variables that describe the surgical task status and interactions among the surgical instruments, objects, and anatomical structures in the physical environment~\cite{yasar2019context, COMPASS}. The definition of surgical context is similar to the tool-tissue interactions (TTI) in action triplets, defined as fine-grained activities in surgical process modeling, which consist of an action verb, a surgical instrument, and the target anatomy~\cite{nwoye2020recognition}. In the CholecTriplet2021 benchmark challenge
for action triplet recognition from surgical videos \cite{nwoye2023cholectriplet2021},
several competing deep-learning methods were developed, including transformer-based with self-attention approaches (e.g., Rendezvous \cite{nwoye2022rendezvous} and SIRNet \cite{li2022sirnet}), convolutional LSTMs, and multi-task learning. In this work, we use context 
definitions for dry-lab surgical tasks from \cite{COMPASS} to perform rule-based context inference based on the masks generated by an STCN model. Similar to previous works mentioned above, we only use video data for context inference because labeled video data is more accessible than robot kinematic data.
\section{Methods}
 Figure \ref{fig:pipeline} shows our overall pipeline for runtime surgical scene segmentation and context inference.
\subsection{Problem Statement}
We have a sequence of input frames $I$ to segment and the first image/mask pair $I_{init} \& M_{init}$. For each frame $I_i$ of size $H \times W $, the memory network's task is to assign a label $m_{hw} \in \{0,1\}$ to each pixel in the image to indicate if pixel $(h,w)$ belongs to an object mask $O$. In our case, we identify the object classes that are important for context inference, $O$=\{ left grasper, right grasper, needle, thread, ring\}.  With the same model but different masks of different objects in the first image/mask pair, multiple networks can be initialized to run in parallel to perform binary segmentation for each object. Through aggregating the binary outputs of all object models, we obtain the segmentation $M$ for each frame in our input images. Then we use the segmented frames to generate surgical context $T$, which is defined as a set of state variables $S_1, S_2, S_3, S_4, S_5$, each describing the status of a task and interactions among the surgical tools and objects in the
physical environment \cite{COMPASS}. As shown in Figure \ref{fig:pipeline}, the first four state variables are used to describe the objects that are being held or are in contact with surgical instruments and are applicable to all tasks. The fifth state variable is specific to the task, such as the position of the needle relative to the fabric or ring in the Suturing and Needle Passing tasks, or the status of the knot in the Knot Tying task.

\subsection{Space-Time Correspondence Network}
Space Time Correspondence Network (STCN) \cite{cheng2021rethinking} takes a set of frames $I$ from a video and the first image/mask pair $I_{init} \& M_{init}$, then proceeds to process the frames one by one while keeping a collection of  key and value in the memory. For every image to segment, the query key is first generated with ResNet50, $E_k(I)$, to obtain the query key $k_Q$. Using the keys $k_M$ encoded from prior frames, which are reused from the previous query keys, an affinity matrix can be obtained that describes the similarities between the current query key $k_Q$ and the memory keys $k_M$. The affinity function, defined as the negative $L2$ similarity, and the normalized affinity matrix $W$ are shown in the following equations: 
\begin{equation}
    S_{ij} = -||k^Q_i - k^M_j||_2^2 
\end{equation}
\begin{equation}
     W_{ij} = \frac{exp(S_{ij})}{\sum_n exp(S_{nj})}
\end{equation}
Value features are generated with an encoder (ResNet18) that takes in both an image and mask $E_v(I, M)$. The memory network can retrieve the corresponding value features $v^Q$ from the previous frames' value features $v^M$ in the memory bank by matrix multiplication as shown in Equation \ref{eq: value}:
\begin{equation}
    v^Q = v^MW
    \label{eq: value}
\end{equation}
Then we can obtain the mask with a decoder $D$ that takes in value features from the matrix multiplication. We follow the same setting as the original paper \cite{cheng2021rethinking} where every fifth frame's value is stored in the memory bank. However, we adapt model post-processing to aggregate the binary segmentation masks of different objects. This is unlike the STCN in \cite{cheng2021rethinking} that passes all the binary masks through an aggregation network module to assign an individual class per pixel, which could introduce unnecessary operation time. We also change the input so that the model can process a non-overlapping moving window of frames to enable runtime inference. Since the same model can be used to segment different objects given different first image/mask pairs, there is no temporal dependency between segmenting the masks of different objects. We can have multiple models running in parallel to segment different objects.    

\subsection{Context Inference}
\label{sec:context inf}
The tool and object interactions, such as ``Left Grasper holding the Needle" as depicted in Figure \ref{fig:pipeline}, can be detected by analyzing intersections and distances between object masks within a given frame. In this paper, we specifically focus on detecting the five state variables that characterize the surgical context in the dry-lab surgical tasks of Suturing, Needle Passing, and Knot Tying~\cite{COMPASS,ICRA23}. As shown in \ref{fig:pipeline}, the first four states describe what the Left Grasper is holding (S1) or in contact with (S2) and what the Right Grasper is holding (S3) or contacting (S4). These variables can take on values representing interactions with nothing (0), the needle (2), the thread (3), or other objects in the surgical scene. 
The last variable (S5) is task-specific and describes the progress within a particular trial. For example, in Needle Passing and Suturing, the Needle can be ``not touching", ``touching" or ``in" with respect to the canvas or ring. 

We use our previously proposed rule-based method from \cite{ICRA23} for context inference. In this method, first a pre-processing step removes noise around needle and thread  masks. Then the contour extraction step removes rough edges and reduces $M$ to a list of points $p$ as polygons for each object class. We use these simplified polygons to calculate intersections and distances between objects for each frame. We drop polygons with areas under 15 pixels to remove segmentation artifacts and smooth the polygons using the Ramer–Douglas–Peucker (RDP) algorithm \cite{RAMER1972244,douglas1973algorithms}.

\setlength{\abovedisplayskip}{-3pt}
\setlength{\belowdisplayskip}{-3pt}
\small
\begin{align} 
\text{Left Hold} &
\begin{cases} \label{equn:LH}
    2   &   \text{if } D(LG,N)<1 \wedge \neg \alpha \\ 
    3   &   \text{if } Inter(LG,T)>0 \wedge \neg \alpha \\
    0   &   \text{otherwise}
\end{cases}
\\
\text{Left Contact} &
\begin{cases} \label{equn:LC}
    2   &   \text{if } D(LG,N)<1 \wedge \alpha \\
    3   &   \text{if } Inter(LG,T)>0 \wedge \alpha \\
    0   &   \text{otherwise} 
\end{cases}
\\
\text{Needle} &
\begin{cases} \label{equn:N}
    2   &   \text{if} (Inter(Ts,N) > 0 \wedge N.x < Ts.x)  \\
    1   &   \text{if} (Inter(Ts,N) = 0 \vee N.x\geq Ts.x) \wedge  \\
        &   (D(RG,T)>1 \vee D(LG,N)>1) \\  
    0   &   \text{otherwise} 
\end{cases}
\end{align}
\normalsize
\setlength{\abovedisplayskip}{6pt}
\setlength{\belowdisplayskip}{0pt}

Overlap between masks is detected by calculating a feature vector $v$ of distances and intersection areas between pairs of input masks, including Left Grasper $(LG)$, Right Grasper $(RG)$, Thread $(T)$, Needle $(N)$, Tissue Points $(Ts)$, and Rings $(R)$. 
The distance and intersection functions $D(I, J)$ and $Inter(I,J)$ are defined as the pixel distance and area of intersection between two object masks $I$ and $J$. Specifically, for any object polygon $I$ which is comprised of several polygon segments $i_1, i_2, ..., i_n$, the distance to any other object $J$ can be calculated as: $D(I, J) = \text{average}([d (i,j) \text{ for } i \in  I \text{ and }  j \in  J])$. The $Inter(I,J)$ function uses a geometric intersection algorithm from the Shapely library \cite{shapely} to calculate the intersection between two object masks. We use $I.x,I.y$ for an object $I$ as the horizontal and vertical coordinates of the midpoint of its polygon $I$, calculated as the average of every point in $I$. The Tissue Points $(Ts)$ represent the markings on the tissue where the needle makes contact in the Suturing task. To determine the Boolean variable $ (\alpha) $,  representing open or closed status of grasper, we use an experimentally found threshold of 18 pixel units to the distance between the grasper jaw ends. Logic vectors are constructed with the Equations \ref{equn:LH}-\ref{equn:N}, which can then be used to estimate the values of state variables. These equations are for the left hand in the Suturing task. Similar sets of equations are used for the right hand and for the Needle Passing  and Knot Tying tasks. 

\section{Experimental Evaluation}
\subsection{Experimental Setup}
We evaluate our proposed approach on a 80/20 train/test split of the JIGSAWS dataset ~\cite{gao2014jhu} in comparison to the state-of-the-art surgical scene segmentation models and a baseline Deeplab V3 model \cite{ICRA23}. The Deeplab model performs binary segmentation by classifying each pixel as the background vs. an object class. 
It is the most recent model evaluated for all the objects and tasks in the JIGSAWS dataset in~\cite{ICRA23}. Binary masks for the tools and objects were obtained through manual labeling at 2Hz from the original 30Hz videos (640*480 pixels per frame) and were used to train and test the segmentation models. We use the context labels (obtained at 3Hz) from \cite{COMPASS} to evaluate the context inference performance. 

In our experiments, we used pretrained weights from the STCN model in \cite{cheng2021rethinking} which was trained with the static image datasets and DAVIS \cite{pont20172017} and YouTubeVOS \cite{xu2018large}. We first trained the model with all the data from the Suturing, Needle Passing, and Knot Tying tasks for 500 epochs. Then we fine-tuned individual models for each task (Suturing, Needle Passing and Knot Tying) and evaluate the respective segmentation performance for each task with 200 epochs. The training process follows the main training process in \cite{cheng2021rethinking}. Specifically, a set of three frames  are arranged in chronological order, with the first frame being the ground-truth mask. The second frame is predicted using the first frame as a memory reference. The information of the first and second frames along with their prediction is stored in the memory bank. The third frame is subsequently predicted by utilizing the combination of the first and second frames. During runtime inference, unlike the main training process, the fifth frame is stored in the memory bank.  

The model training was done on a 64-bit PC with an Intel Core i9 CPU @  3.70GHz  and  32GB  RAM running Linux  Ubuntu  20.04 and an NVIDIA RTX2080 Ti 11GB GPU. The standard metric, mean Intersection over Union (IOU), is used to evaluate segmentation and context inference performance \cite{lin2014microsoft}. Each predicted segment or context is matched to a corresponding segment in the ground truth. Then, the average IOU for each class or context state is calculated.

We assume that during deployment the model will segment video data with the same object classes and similar physical context as the ones in the first image/mask pair from training. 
Several STCN models corresponding to different object classes can run in a parallel to segment different objects.

\subsection{Different First Image/Mask Pairs}
\label{sec: init pairs}
To understand the effect of the first image/mask pairs on the network performance, we experimented with various ways of generating first image/mask pairs. Specifically, we examined the STCN model performance in the following setups for the first frame/mask (FF) pair: DeepLab-FF, Train-FF and GT-FF. In the Deeplab-FF setup, we used the image and the corresponding mask generated by the baseline Deeplab model from the frame that an object first appears. In the Train-FF setup, image/ground-truth mask pairs are from the training set. To create the sets of image/mask pairs for the Train-FF setup, we visually inspected the image to ensure that the appearance of the object is representative of the object in the training set. In the GT-FF setup, similar to setup in ~\cite{cheng2021rethinking} the first image/mask pairs are from the first ground-truth mask in the test set. Note that in real deployment settings for runtime segmentation ground-truth masks for test data will not be available. We only used the GT-FF setup for performance comparison. These preliminary experiments showed that the Train-FF consistently achieves the same (for graspers) or slightly better (for Needle and Thread) performance than the other setups and is more suitable for real-world deployment, so we used this setup for our STCN model in the rest of the experiments. Deeplab-FF setup could be also suitable for real-world deployment, in scenarios where the test images might be significantly different from the train images.

\subsection{Context Inference Performance}
To evaluate the effect of the segmentation performance on the performance of the context inference, we fed the ground truth segmentation masks as well as the masks generated by the STCN and the baseline Deeplab models from Section \ref{sec: init pairs} to the context inference component. The context inference performance from the ground truth segmentation masks can serve as a baseline for the max possible performance we can get from the rule-based approach. We calculated the average IOU for each context state and compared the results with respect to the ground truth context labels from \cite{COMPASS}. 

\subsection{Batched Performance \& Time Analysis}
The original memory network was designed to perform offline and process the whole video at once. However, to perform runtime inference, the memory network should be able to process a sequence of frames within the video timing constraints without significant performance degradation. 
In this experiment, we evaluated the runtime performance of the network by analyzing the computation time for segmentation and context inference given different batch sizes. Based on our analysis, the smallest interval between a change of context in the JIGSAWS videos is about 333 milliseconds, which corresponds to a batch size of 10 given a 30Hz video. We explored different batch sizes, ranging from 5 to 25 frames per batch and examined the time taken to perform the segmentation along with the context inference for the last frame in the batch. 
The context changes in between the frames are ignored because within each batch the duration between consecutive images can be within milliseconds, which is too short for context changes to happen. We kept the duration between each batch within 1 second to ensure we can capture the network behavior when the STCN memory bank only stores the key, value pairs of a few prior frames, which in this case, 1, 2, 3, 4, and 5 prior frames' information are stored in the memory bank, respectively. 


\section{Results and Discussion}
\subsection{Comparison to the State of the Art}
Our results in Table \ref{tab:per-comp-sota} show that STCN with the image/ground-truth mask pairs from the training set (Train-FF) have better performance than the baseline Deeplab.
We can see that STCN has over 20\% performance improvement for the left grasper, right grasper, needle, thread and ring across all the three tasks of Suturing, Needle Passing and Knot Tying. There is in particular significant improvement for  segmentation of more difficult object classes, including needle, thread and ring. We observe over 200\% IOU improvements for the needle class in the Suturing and Needle Passing tasks. There are similar improvements for the thread and ring classes. The needle class has more performance improvement than the thread class, but the overall IOU for the needle class in Suturing and Needle Passing tasks (0.57 and 0.32) is still lower than the thread (0.83 and 0.58) class. This is because the needle masks are generally smaller than the thread masks, so any inaccuracies in the masks for the needle results in a lower IOU than the thread.  

\begin{table}[t!]
\vspace{0.2cm}
\centering
\caption{Tool and Object Segmentation Performance (Mean IOU per Object Class) for the MICCAI Endovis 18 (M) and JIGSAWS Suturing (S), Needle Passing
(NP), and Knot Tying (KT) tasks.}
\label{tab:per-comp-sota}
\resizebox{\columnwidth}{!}{%
\begin{tabular}{|c|c|cc|ccc|c}
\cline{1-7}
\multirow{2}{*}{Model}                                                           & \multirow{2}{*}{Data}   & \multicolumn{2}{c|}{Graspers}      & \multicolumn{3}{c|}{Objects}                                      &  \\ \cline{3-7}
                                                                                 &                         & \multicolumn{1}{c|}{Left}  & Right & \multicolumn{1}{c|}{Needle} & \multicolumn{1}{c|}{Thread} & Ring  &  \\ \cline{1-7}
DeepLab v3+ \cite{allan20202018}                                                                    & \multirow{2}{*}{M} & \multicolumn{2}{c|}{0.78}          & \multicolumn{1}{c|}{0.014}  & \multicolumn{1}{c|}{0.48}   & -   &  \\ \cline{1-1} \cline{3-7}
U-net   \cite{allan20202018}                                                                          &                         & \multicolumn{2}{c|}{0.72}          & \multicolumn{1}{c|}{0.02}   & \multicolumn{1}{c|}{0.33}   & -   &  \\ \cline{1-7}
Mobile-U-Net  \cite{andersen2021real}                                                                   & S                       & \multicolumn{2}{c|}{0.69}          & \multicolumn{1}{c|}{0.56}   & \multicolumn{1}{c|}{-}    & -  &  \\ \cline{1-7}
U-Net \cite{papp2022surgical}                                                                           & S                       & \multicolumn{2}{c|}{0.66}          & \multicolumn{1}{c|}{-}    & \multicolumn{1}{c|}{-}    & -   &  \\ \cline{1-7}
LinkNet  \cite{papp2022surgical}                                                                        & S                       & \multicolumn{2}{c|}{0.80}          & \multicolumn{1}{c|}{-}    & \multicolumn{1}{c|}{-}    & -   &  \\ \cline{1-7}
\multirow{3}{*}{\begin{tabular}[c]{@{}c@{}}Baseline (Deeplab)\end{tabular}} & S                       & \multicolumn{1}{c|}{0.71}  & 0.64  & \multicolumn{1}{c|}{0.19}   & \multicolumn{1}{c|}{0.52}   & -   &  \\ \cline{2-7}
                                                                                 & NP                      & \multicolumn{1}{c|}{0.61}  & 0.49  & \multicolumn{1}{c|}{0.09}   & \multicolumn{1}{c|}{0.25}   & 0.37  &  \\ \cline{2-7}
                                                                                 & KT                      & \multicolumn{1}{c|}{0.74}  & 0.61  & \multicolumn{1}{c|}{-}    & \multicolumn{1}{c|}{0.44}   & -   &  \\ \cline{1-7}
\multirow{3}{*}{STCN (Train-FF)}                                                     & S                       & \multicolumn{1}{c|}{\textbf{0.88}} & \textbf{0.84} & \multicolumn{1}{c|}{\textbf{0.57}}  & \multicolumn{1}{c|}{\textbf{0.83}}  & -   &  \\ \cline{2-7}
                                                                                 & NP                      & \multicolumn{1}{c|}{0.83} & 0.77 & \multicolumn{1}{c|}{0.32}  & \multicolumn{1}{c|}{0.58}  & \textbf{0.66} &  \\ \cline{2-7}
                                                                                 & KT                      & \multicolumn{1}{c|}{0.84} & 0.82 & \multicolumn{1}{c|}{-}    & \multicolumn{1}{c|}{0.79}  & -   &  \\ \cline{1-7}
\end{tabular}%
}
\vspace{-1.5em}
\end{table}

We also compare the performance of our method with the state-of-the-art surgical scene segmentation models developed using the MICCI Endovis 2018 dataset \cite{allan20202018} and the JIGSAWS dataset \cite{gao2014jhu}. The Endovis 2018  does not have the left and right grasper class, so we use the clasper class that has the closest resemblance to the JIGSAWS's graspers. Although the Endovis 2018  does not differentiate left and right graspers, we compare the performance of our left and right grasper classes with the single clasper class in the Endovis 2018. 
The graspers, needle, and thread performance are better in our model in the the Suturing, Needle Passing, and Knot Tying tasks in comparison to the Deeplab v3 as well as U-net in the Endovis 2018. For the JIGSAWS dataset, there are no publicly available segmentation labels for the objects in the video. The Mobile-U-net in \cite{andersen2021real} uses the Suturing segmentation labels annotated by the authors and does not differentiate between left and right grasper classes. Our network achieves better performance in the grasper class (0.88/0.84 vs. 0.69) and comparable performance for the needle class in the Suturing task (0.57 vs. 0.56). However, the Mobile-U-net has not been evaluated on the thread class in the Suturing task and has not been evaluated on the Needle Passing and Knot Tying tasks. We should note that the Mobile-U-net and our network are not evaluated on the same set of images from the Suturing videos, and there could be discrepancies in the labels. One recent work \cite{papp2022surgical} trained different networks to perform segmentation on the left/right graspers along with the shaft using labels generated from an optical flow method for the Suturing task. Here we show two of their top performing networks, including the UNet (0.66) and the LinkNet (0.80). Our method also has better performance than these two networks. 

\begin{figure*}[ht!]
\vspace{0.2cm}
    \centering
    \begin{subfigure}{0.2\textwidth}
        \centering
        \includegraphics[width=\textwidth]{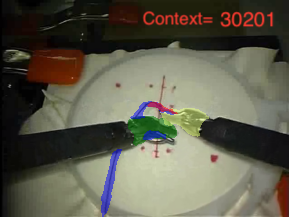}
        \caption{Deeplab Baseline}
        \label{fig:dl-b}
    \end{subfigure}
    \begin{subfigure}{0.2\textwidth}
        \centering
        \includegraphics[width=\textwidth]{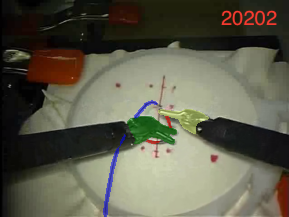}
        \caption{STCN(Train-FF)}
        \label{fig:stcn}
    \end{subfigure}
    \begin{subfigure}{0.2\textwidth}
        \centering
        \includegraphics[width=\textwidth]{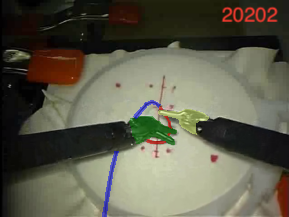}
        \caption{Ground Truth}
        \label{fig:gt}
    \end{subfigure}
    \vspace{-0.5em}
    \caption{Comparison of Deeplab Baseline to STCN Outputs}
    \label{fig:dl-full}
    \vspace{-0.5em}
\end{figure*}

\subsection{Context Inference}
\label{sec:d-context}
In this section, we compare the results of context inference given the segmentation masks in Table \ref{tab:context-infer} and the ground-truth masks. In task Suturing, we observe that for Left Contact and Right Contact states, we achieve slightly better performance with the STCN model than the baseline Deeplab. The contact states involve logic to calculate the distance and intersection between the grasper masks and the needle and thread masks. Since our model generates segmentation masks that are closer to the ground truth, we see an improvement in detecting these states.  The most obvious improvement is in detecting the needle state (0.083). This is expected because we have the needle's performance improved by $\sim$200\% in comparison to the baseline Deeplab. In other tasks, we see a similar trend as Suturing. In Needle Passing, we see that our model has comparable performance in detecting Left Hold, Left Contact, and Right Contact states as the baseline Deeplab and the ground truth and achieves better performance for the Needle State. In task Knot Tying, our model achieves comparable or better performance than the baseline Deeplab across all five states.

\begin{table}[b!]
\centering
\vspace{-1.75em}
\caption{Context Inference Performance }
\vspace{-0.5em}
\label{tab:context-infer}
\resizebox{\columnwidth}{!}{%
\begin{tabular}{|l|l|l|l|l|l|l|l|}
\hline
Tasks                                                                    & Setup                                                        & \begin{tabular}[c]{@{}l@{}}Left \\ Hold\end{tabular}        & \begin{tabular}[c]{@{}l@{}}Left \\ Contact\end{tabular}     & \begin{tabular}[c]{@{}l@{}}Right \\ Hold\end{tabular}     &\begin{tabular}[c]{@{}l@{}}Right \\ Contact\end{tabular}      &  \begin{tabular}[c]{@{}l@{}}Needle/ \\ Knot\end{tabular}      & Avg.        \\ \hline
\multirow{3}{*}{Suturing}                                                 & \begin{tabular}[c]{@{}l@{}}STCN\\ (Train-FF)\end{tabular}    & 0.417          & \textbf{0.774} & 0.561          & \textbf{0.869} & \textbf{0.383} & \textbf{0.601} \\ \cline{2-8} 
                                                                          & \begin{tabular}[c]{@{}l@{}}Baseline\\ (Deeplab)\end{tabular} & \textbf{0.478} & 0.751          & \textbf{0.603} & 0.866          & 0.3            & 0.6            \\ \cline{2-8} 
                                                                          & Ground Truth                                                 & 0.524          & 0.765          & 0.605          & 0.869          & 0.388          & 0.63           \\ \hline
\multirow{3}{*}{\begin{tabular}[c]{@{}l@{}}Needle\\ Passing\end{tabular}} & \begin{tabular}[c]{@{}l@{}}STCN\\ (Train-FF)\end{tabular}    & 0.374          & \textbf{0.967} & 0.259          & 0.939          & \textbf{0.416} & \textbf{0.577} \\ \cline{2-8} 
                                                                          & \begin{tabular}[c]{@{}l@{}}Baseline\\ (Deeplab)\end{tabular} & \textbf{0.398} & 0.967          & \textbf{0.658} & \textbf{0.946} & 0.393          & \textbf{0.577} \\ \cline{2-8} 
                                                                          & Ground Truth                                                 & 0.415          & 0.968          & 0.648          & 0.942          & 0.411          & 0.586          \\ \hline
\multirow{3}{*}{\begin{tabular}[c]{@{}l@{}}Knot \\ Tying\end{tabular}}    & \begin{tabular}[c]{@{}l@{}}STCN\\ (Train FF)\end{tabular}    & \textbf{0.778} & \textbf{0.782} & \textbf{0.597} & \textbf{0.801} & 0.582          & \textbf{0.708} \\ \cline{2-8} 
                                                                          & \begin{tabular}[c]{@{}l@{}}Baseline\\ (Deeplab)\end{tabular} & 0.746          & 0.724          & 0.571          & 0.783          & \textbf{0.588} & 0.682          \\ \cline{2-8} 
                                                                          & Ground Truth                                                 & 0.825          & 0.766          & 0.606          & 0.791          & 0.619          & 0.721         \\ \hline
\end{tabular}%
}
\end{table}

\begin{table*}[h!]
\centering
\caption{95\% Confidence Interval (CI) and Standard Deviation for IOU for Batched Input }
\vspace{-0.5em}
\label{tab:conf_interval}
\resizebox{\textwidth}{!}{%
\begin{tabular}{|c|ccc|ccc|ccc|ccc|ccc|}
\hline
IOU & \multicolumn{3}{c|}{Left Grasper}                                & \multicolumn{3}{c|}{Right Grasper}                               & \multicolumn{3}{c|}{Needle}                                      & \multicolumn{3}{c|}{Thread}                                      & \multicolumn{3}{c|}{Ring}                                        \\ \hline
                   & \multicolumn{2}{c|}{CI  (-/+)}                                 & std    & \multicolumn{2}{c|}{CI  (-/+)}                                 & std    & \multicolumn{2}{c|}{CI  (-/+)}                                 & std    & \multicolumn{2}{c|}{CI  (-/+)}                                 & std    & \multicolumn{2}{c|}{CI  (-/+)}                                 & std    \\ \hline
Suturing           & \multicolumn{1}{c|}{0.874} & \multicolumn{1}{c|}{0.879} & 0.0015 & \multicolumn{1}{c|}{0.840} & \multicolumn{1}{r|}{0.846} & 0.0017 & \multicolumn{1}{r|}{0.532} & \multicolumn{1}{r|}{0.552} & 0.0061 & \multicolumn{1}{r|}{0.812} & \multicolumn{1}{r|}{0.825} & 0.0037 & \multicolumn{3}{l|}{}                                            \\ \hline
Needle Passing     & \multicolumn{1}{r|}{0.825} & \multicolumn{1}{r|}{0.828} & 0.001 & \multicolumn{1}{r|}{0.712} & \multicolumn{1}{r|}{0.742} & 0.0093 & \multicolumn{1}{r|}{0.228} & \multicolumn{1}{r|}{0.292} & 0.0197 & \multicolumn{1}{r|}{0.560} & \multicolumn{1}{r|}{0.585} & 0.0077 & \multicolumn{1}{r|}{0.648} & \multicolumn{1}{r|}{0.665} & 0.005 \\ \hline
Knot Tying         & \multicolumn{1}{r|}{0.834} & \multicolumn{1}{r|}{0.842} & 0.0024 & \multicolumn{1}{r|}{0.815} & \multicolumn{1}{r|}{0.823} & 0.0025 & \multicolumn{3}{l|}{}                                            & \multicolumn{1}{r|}{0.787} & \multicolumn{1}{r|}{0.799} & 0.0038 & \multicolumn{3}{l|}{}                                            \\ \hline
\end{tabular}%
}
\vspace{-1.75em}
\end{table*}
To provide an illustrative example, Figure \ref{fig:dl-full} shows the segmentation masks and inferred context from Deeplab, STCN, and ground-truth. We see that the STCN segmentation mask in \ref{fig:stcn} can segment the lower part of the needle which the baseline Deeplab in \ref{fig:dl-b} misses. Therefore, the STCN mask helps to infer the Left Hold - Needle state correctly ($  D(LG,N)<1 \wedge \neg \alpha=True$), which is not the case for the baseline Deeplab mask. 
Rather, the lower part of the grasper for the baseline Deeplab mask is falsely segmented to be the thread. As a result, the context is generated incorrectly ($Inter(LG,T)>0 \wedge \neg \alpha=True $) as Left Hold - Thread. The needle state is also inferred incorrectly for the baseline Deeplab mask, which the STCN segmentation mask corrects. 

For the Hold states in Suturing and Needle Passing, the baseline Deeplab achieves better context inference performance, even though we have more accurate masks for the left and right graspers, needle, and thread.  
 In the rules for detecting these states, the distance and intersection are evaluated based on whether the distance is smaller than a specific threshold $ D(LG,N)<1$ and if there is an intersection between the masks $Inter(LG,T)>0$. The thresholds were selected based on the ground truth in the training set, which could be biased against mask outputs from the model, so having hard thresholds may not be appropriate. Although the rules for the Contact states also rely on the distance and intersection logic, they are less affected by this problem. 

\subsection{Batched Performance \& Time Analysis}
Table \ref{tab:conf_interval} presents the 90\% confidence intervals for the IOU performance with the batched inputs of 5 to 25 frames. We see that the batched performance is approximately the same as the IOU performance of the offline model that processes the video of the whole trial all at once. However, the Suturing needle class, the Knot Tying right grasper, and the needle class have slightly lower performance than the offline model. Because batched input reduces the number of image and mask pairs encoded in the memory bank, the performance can be reduced especially the object already has less presence in the train set such as the needle class for the Suturing and Needle Passing tasks. The right grasper class of the Needle Passing task also has less performance than the offline model. This could be due to the Needle Passing task having lower-quality videos. However, despite having a smaller memory bank, our batched performance is still better than the baseline Deeplab model in Table \ref{tab:per-comp-sota}. We also observe that using batched input results in a small variance between the performance of different input batched sizes. This means that our method's performance can be robust when processing batched inputs.   


In Table \ref{tb:context-time}, we show the time to perform segmentation for different batch sizes and the time to perform context inference for the last frame. We observe an approximately linear increase in the total time to perform both segmentation and context inference. In the JIGSAWS dataset, the video is captured at 30Hz, so the batch sizes of 5, 10, 15, 20, and 25 correspond to 167, 333, 500, 667, and 833 milliseconds of data, respectively. A batch size of 10 is the best for runtime context inference because it has more frames stored in the memory and can capture the smallest change in context. We observe that the segmentation and context inference can be efficiently completed within the runtime constraints. For example, for a batch size of 10, when each batch arrives, we have 333 ms to process the whole batch before the next batch arrives. We see that the segmentation and context inference take, respectively, total of 205, 207 201 ms to complete for the Suturing, Needle Passing and Knot Tying tasks, which are well within the total time window of 333 ms.  

\begin{table}[h!]
\vspace{-0.5em}
\caption{Segmentation and Context Inference Time per Batch}
\vspace{-0.75em}
\resizebox{\columnwidth}{!}{%
\begin{tabular}{|c|c|cc|cc|cc|}
\hline
\multirow{2}{*}{\begin{tabular}[c]{@{}c@{}}Batch Size\\ (Time ms)\end{tabular}} & \multirow{2}{*}{\begin{tabular}[c]{@{}c@{}}All Tasks\\ Segmentation\end{tabular}} & \multicolumn{2}{c|}{Suturing}        & \multicolumn{2}{c|}{Needle Passing}  & \multicolumn{2}{c|}{Knot Tying}      \\ \cline{3-8} 
                                                                             &                                                                                   & \multicolumn{1}{c|}{Context} & Total & \multicolumn{1}{c|}{Context} & Total & \multicolumn{1}{c|}{Context} & Total \\ \hline
5 (167 ms)                                                                   & 82                                                                                & \multicolumn{1}{c|}{29}      & 111   & \multicolumn{1}{c|}{46}      & 128   & \multicolumn{1}{c|}{37}      & 119   \\ \hline
10 (333 ms)                                                                  & 180                                                                               & \multicolumn{1}{c|}{26}      & 206   & \multicolumn{1}{c|}{27}      & 207   & \multicolumn{1}{c|}{21}      & 201   \\ \hline
15 (500 ms)                                                                  & 286                                                                               & \multicolumn{1}{c|}{22}      & 308   & \multicolumn{1}{c|}{26}      & 312   & \multicolumn{1}{c|}{19}      & 305   \\ \hline
20 (667 ms)                                                                  & 424                                                                               & \multicolumn{1}{c|}{22}      & 446   & \multicolumn{1}{c|}{25}      & 448   & \multicolumn{1}{c|}{17}      & 441   \\ \hline
25 (833 ms)                                                                  & 576                                                                               & \multicolumn{1}{c|}{21}      & 597   & \multicolumn{1}{c|}{24}      & 600   & \multicolumn{1}{c|}{18}      & 594   \\ \hline
\end{tabular}%
}

\label{tb:context-time}
\vspace{-1em}
\end{table}


\section{Conclusion}
In this work, we improve the current state-of-the-art surgical scene segmentation with an STCN memory network to better segment difficult objects (e.g., needle and thread) and provide temporal consistency for the masks. Our experiments using data from dry-lab simulation tasks demonstrate that the STCN model can achieve superior performance compared to several baselines and can process smaller batches of data at runtime with minimal impact on performance. We also show that the STCN model does not necessarily require an image/mask pair from the first frame of the video. Instead, selecting a frame that represents the object's appearance in the training set can lead to similar or better performance. 
Further, the improved segmentation performance has positive influence on context inference, particularly detection of needle and thread states. Our time analysis confirms that both the segmentation and context inference can be performed within the runtime constraints, opening up possibilities for runtime applications like surgical workflow analysis, skill assessment, and error detection. Future work will focus on evaluating this method using data from real surgical procedures. 





\section*{ACKNOWLEDGMENT}
This work was supported in part by the National Science Foundation grant CNS-2146295.


\bibliographystyle{IEEEtran}
\bibliography{irosbib}
\end{document}